\documentclass[letterpaper, 10 pt, conference]{ieeeconf}  

\IEEEoverridecommandlockouts                              

\usepackage{graphics} 
\usepackage{epsfig} 
\usepackage{mathptmx} 
\usepackage{times} 
\usepackage{amsmath} 
\usepackage{amssymb}  
\usepackage{algorithm}
\usepackage{algorithmic}
\usepackage{xcolor}
 \usepackage[subrefformat=parens]{subcaption}

\usepackage{multirow}
\usepackage{cite}
\usepackage{dsfont}

\usepackage{calrsfs}
\DeclareMathAlphabet{\pazocal}{OMS}{zplm}{m}{n}

\title{\LARGE \bf
TransTouch: Learning Transparent Objects Depth Sensing Through Sparse Touches
}

\author{Liuyu Bian$^{*1}$, Pengyang Shi$^{*1}$, Weihang Chen$^{2}$, Jing Xu$^{2}$, Li Yi$^{+1,3,4}$, and Rui Chen$^{+2}$ 
\thanks{$^*$ indicates equal contribution; $^+$ indicates equal advising.}
\thanks{Correspondences: {\tt \small ericyi@mail.tsinghua.edu.cn}, {\tt \small chenruithu@mail.tsinghua.edu.cn.}}
\thanks{$^{1}$The authors are with Institute for Interdisciplinary Information Sciences, Tsinghua University, Beijing, China, 100084.}
\thanks{$^{2}$The authors are with Department of Mechanical Engineering, Tsinghua University, Beijing, China, 100084.}
\thanks{$^{3}$The author is with Shanghai Artificial Intelligence Laboratory, China.}
\thanks{$^{4}$The author is with Shanghai Qi Zhi Institute, China.}
}

\begin{document}

\maketitle
\thispagestyle{empty}
\pagestyle{empty}

\begin{abstract}

Transparent objects are common in daily life. However, depth sensing for transparent objects remains a challenging problem. While learning-based methods can leverage shape priors to improve the sensing quality, the labor-intensive data collection in real world and the sim-to-real domain gap restrict these methods' scalability. In this paper, we propose a method to finetune a stereo network with sparse depth labels automatically collected using a probing system with tactile feedback. We present a novel utility function to evaluate the benefit of touches. By approximating and optimizing the utility function, we can optimize the probing locations given a fixed touching budget to better improve the network's performance on real objects. We further combine tactile depth supervision with a confidence-based regularization to prevent over-fitting during finetuning. 
To evaluate the effectiveness of our method, we construct a real-world dataset including both diffuse and transparent objects. Experimental results on this dataset show that our method can significantly improve real-world depth sensing accuracy, especially for transparent objects.

\end{abstract}

\section{INTRODUCTION}
Allowing active stereo cameras to accurately sense the depth of transparent objects has gained wide attention in robotics~\cite{sajjan2020clear,jiang2022shall, hong2022cluedepth} due to its applications in grasping and manipulation. It has been a long-standing challenge though since the complex light path violates the assumptions made in classic stereo-matching algorithms.

\begin{figure}[ht]
    \centering
    \includegraphics[width=8cm]{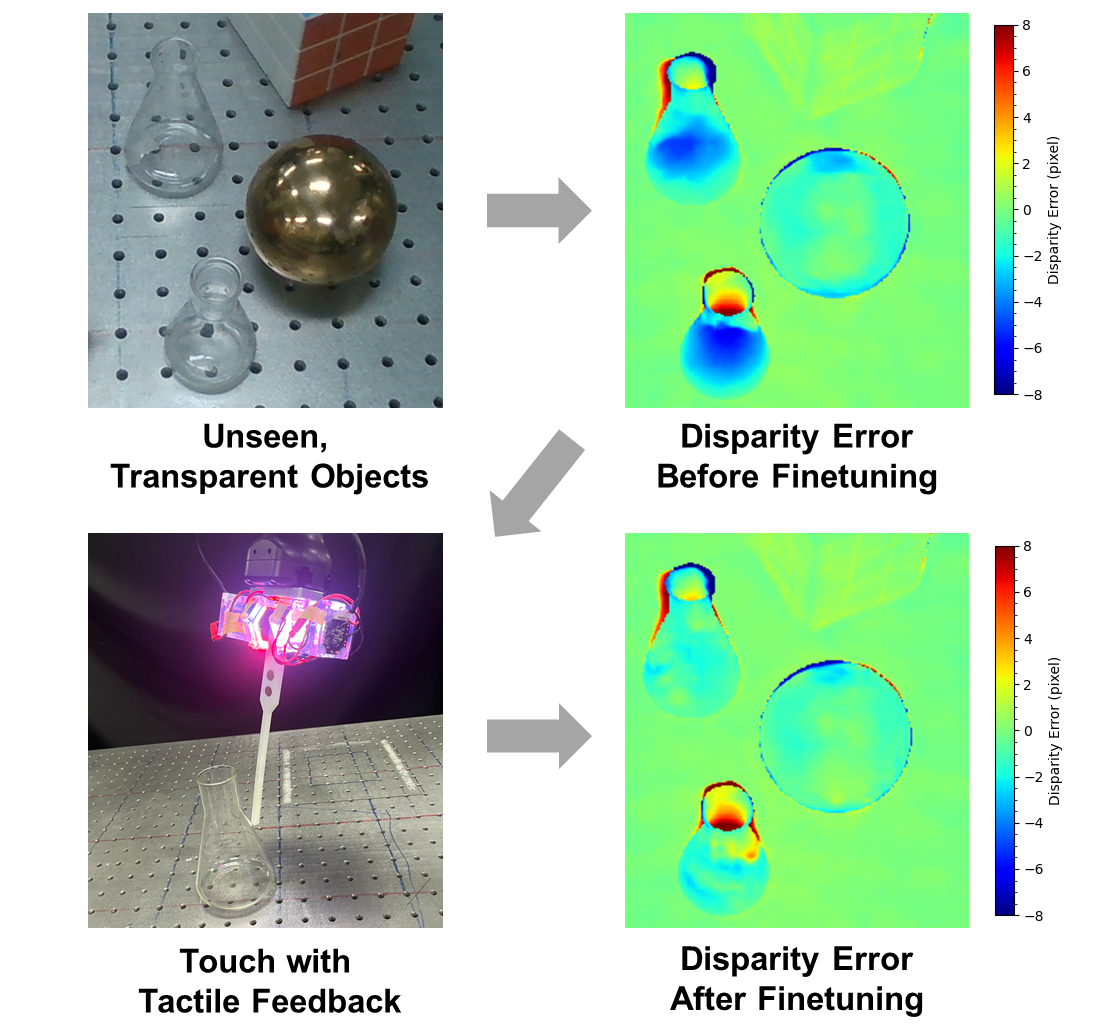}
    \caption{Our work improves the stereo network's performance on unseen transparent objects by using the sparse depth labels that we obtain automatically from touching with tactile feedback.}
    \label{fig:teasor}
    \vspace*{-1.0\baselineskip}
\end{figure}

To tackle this challenge, researchers have turned to deep neural networks where prior samples can be leveraged to understand how to handle challenging stereo-matching cases. Training such a network requires both stereo observations and the corresponding ground truth depth. A synthetic dataset is usually a good start since ground truth depth can be easily obtained for transparent objects. But it is not sufficient due to its domain gap with real-world stereo observations. Usually, some real data is needed to further finetune the model. To cope with the data issue, existing works either collect dense real-world depth labels for transparent objects~\cite{fang2022transcg} or completely give up collecting real depth labels while exploiting self-supervision signals or domain randomization to mitigate the sim-to-real gap~\cite{liu2022activezero, dai2022domain}. The former approach requires a labor-intensive capturing system where every real transparent object needs to be 3D modeled manually, leading to severe data scalability issues. While the latter approach suffers from its restricted performance due to the lack of real-world depth labels, preventing its use in real applications.

Instead of relying on dense real-world depth labels or completely ignoring them, our key idea is to collect sparse real-world depth labels through a selective tactile probing system. As shown in Fig.~\ref{fig:teasor}, we equip a robot arm with a tactile sensor to selectively touch patches in a scene and read out the local depth information in each region it touches. We use the sparse depth supervision with regularization to finetune a pre-trained stereo network, whose performance on unseen transparent objects is improved after finetuning.
Our system is truly automatic with minimal human interventions. It is much more labor-efficient than acquiring dense labels and easier to scale up since we do not need to manually 3D model each real-world transparent object beforehand. While at the same time, the collected sparse label can still successfully improve the network's performance on transparent objects due to its selective nature.

However, we are faced with two technical challenges: 1) how to select the pixels to touch given a fixed amount of touching budget; 2) how to finetune the network with only a sparse set of depth labels. To tackle the first challenge, we present a novel utility function to approximate the effectiveness of touching a certain set of pixels. The utility function is based on the dynamic behavior of a network and can effectively exclude redundant touches. And we also develop a greedy strategy to optimize the utility function. Regarding the second challenge, we propose a novel finetuning strategy dilating sparse depth labels and mixing them with regularization.
To verify the effectiveness of the above novel designs, we create an improved version of the real dataset used in~\cite{liu2022activezero} and conduct comparisons and ablations. Experiments show that our selective sparse depth probing strategy can reduce the depth sensing error of transparent objects from 17.5 mm to 15.5 mm after finetuning with only 100 touch.

To summarize, our contributions are four folds: 1) we design a novel visual-tactile depth probing system to collect sparse depth labels to improve depth sensing on transparent objects; 2) we develop a utility function as well as a utility optimization scheme for touching pixel choosing; 3) we present a finetuning strategy with the sparse depth labels which allows a pre-trained stereo network to better estimate the depth of novel transparent objects; 4) we build a real-world transparent object active stereo dataset based upon~\cite{zhang2023close}, which we will make available to the research community.

\begin{figure*}[t]
    \centering
    \includegraphics[width=17.5cm]{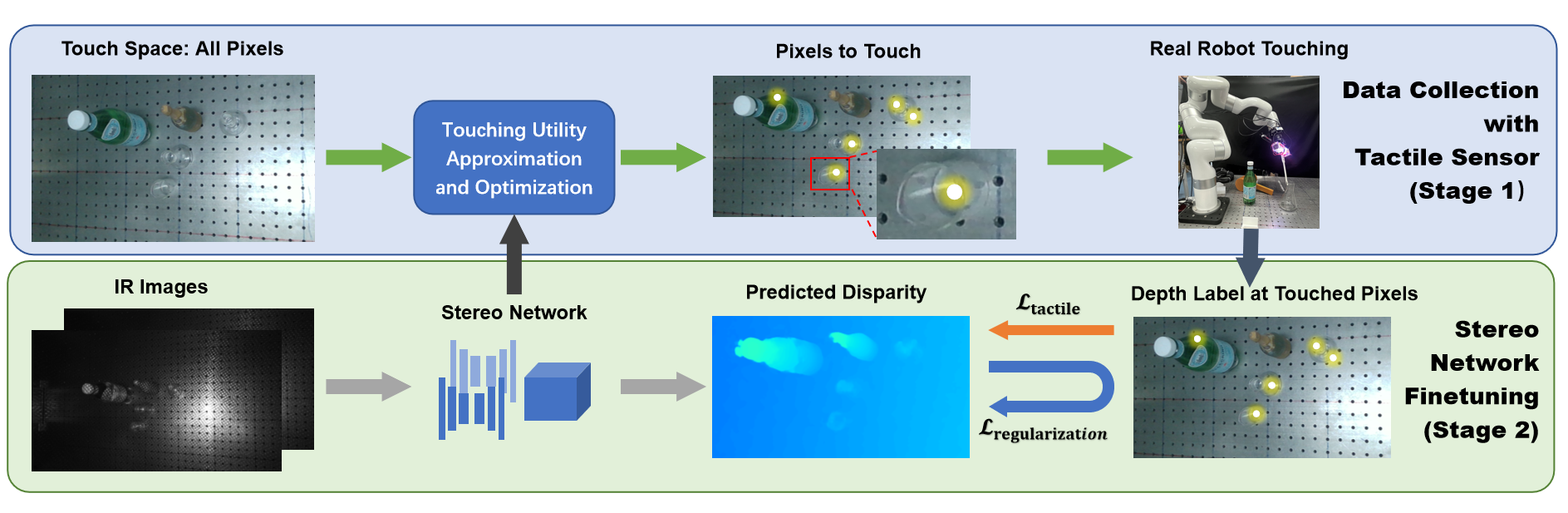}
    \caption{Framework overview. We finetune a pre-trained stereo network with depth label collected with a tactile sensor in a two-stage way. The first stage involves collecting depth labels using a tactile sensor attached to a robot arm. A novel utility function is introduced to enhance the sample efficiency. In the second stage, the pre-trained model is finetuned using the data collected in Stage 1, resulting in improved performance, particularly for transparent objects.
    }
    \label{fig: pipeline}
    \vspace*{-1.0\baselineskip}

\end{figure*}

\section{RELATED WORKS}
\subsection{Learning Based Stereo} 
A traditional pipeline of stereo depth estimation is composed of 4 steps: matching cost computation, cost aggregation, disparity computation, and refinement~\cite{scharstein2002taxonomy}. Learning-based methods gain prevalence in recent years, as they can achieve better results with larger models and computation abilities. Some works replace human-designed features with learned features \cite{zagoruyko2015learning,zbontar2015computing,zbontar2016stereo}. Others train an end-to-end network to directly predict depth from input images. Dispnet\cite{mayer2016large} is the first work to use an end-to-end encoder-decoder model to estimate the disparity. PSMNet\cite{chang2018pyramid} improves feature extraction by introducing a pyramid pooling module to incorporate global context into image features and a stacked hourglass 3D CNN to extend the regional support of context in the cost volume.

However, most learning-based approaches heavily rely on a large-scale dataset with ground truth depth information, which is very labor-intensive to obtain in practice. In ActiveZero~\cite{liu2022activezero}, a training framework that does not require real ground truth depth annotation is proposed, which combines ground truth disparity supervision in the synthetic domain and reprojection self-supervision in the real domain. While it can accurately estimate the disparity on real diffuse and specular objects, its performance on real transparent objects is still inferior. In this paper, we propose a visual-tactile depth probing system that uses a robot with tactile sensors to collect real-world depth and finetune the stereo network to further improve its performance on real transparent objects.

\subsection{Depth Estimation for Transparent Objects}

Transparent objects are common in daily life. However, due to complex light reflection and refraction, current commercial optical depth sensors and stereo-matching methods perform poorly on these objects, limiting their application for downstream tasks, such as grasping and manipulation. ClearGrasp~\cite{sajjan2020clear} first predicts a transparent object mask on the RGB image, then uses global optimization to  complete the depth from the commercial depth sensor with additional surface normal and boundary constraints. However, global optimization is time-consuming for real-world tasks, and it is highly sensitive to normal prediction. 
LIDF~\cite{zhu2021rgb} introduces an implicit representation defined on ray-voxel pairs and refines depth iteratively, but it suffers from the discontinuity of voxelization. DREDS~\cite{dai2022domain} creates a large dataset with realistic sensor noise simulation and domain randomization and trains a depth restoration network to restore the depth of transparent objects, but it suffers from the significant domain gap in images of transparent objects. TransCG~\cite{fang2022transcg} builds a large-scale real-world transparent dataset by tracking IR markers attached to transparent objects. Experimental results demonstrate that real-world depth supervision can significantly improve the network's performance on transparent objects. However, while the data collection procedure can be autonomous, it still needs extensive human labor to build accurate CAD models of real objects. In this paper, we propose a novel visual-tactile depth probing system to acquire ground truth depth of real transparent objects that requires minimum human involvement. And we further propose a touching pixel selection strategy to improve the touch signal's effect on network finetuning through touching utility approximation and optimization. 

\vspace*{-0.05\baselineskip}

\subsection{Vision and Tactile}
The vision signal can provide a global view of the whole scene, but is influenced by the object materials and environmental illumination. In contrast, the tactile signal can provide accurate geometry information of a small area and is robust to material and illumination.

Previous works utilize the complementary nature of the two modalities for 3D shape reconstruction. These works first convert vision and tactile signals into some intermediate representations, such as sparse point clouds\cite{bjorkman2013enhancing,ilonen2014three,gandler2020object}, voxels\cite{wang20183d,watkins2019multi} or meshes\cite{smith20203d,smith2021active}, then merge the two signals and apply refinement on the merged representations. These works focus on reconstructing the object instance's geometry. In contrast, we use the tactile signal to acquire sparse depth labels for stereo network finetuning, and the finetuned network can generalize to unseen transparent objects.

Moreover, the efficiency of touching is highly determined by the selected position. Some works use heuristics, such as random selecting \cite{smith20203d} or always touch areas invisible to camera\cite{watkins2019multi}, which may leave many areas with rich information unexplored. Others use uncertainty of prediction \cite{wang20183d} or a learning-based strategy\cite{rustler2022active,smith2021active} to guide exploration. We explore along this direction and present a novel utility function to improve the selection efficiency.

\section{METHOD}
In this section, we introduce our framework, as shown in Fig.~\ref{fig: pipeline}, for finetuning a pre-trained stereo network with depth label collected with a tactile sensor. The whole framework is combined of two stages. In the first stage (Sec.~\ref{sec:utility_def}), we collect sparse depth labels with a tactile sensor mounted on the robot arm. We present a novel utility function to approximate the effectiveness of pixel-choosing and thus improve sample efficiency. In the second stage (Sec.~\ref{sec:tuning}), a pre-trained model is further finetuned with data collected in Stage 1 with our proposed confidence-based regularization.

The task of our work is to improve the performance of a pre-trained stereo network for real-world images, especially on transparent objects, by using a limited number of tactile probings. The pre-trained stereo network $f$ predicts disparity $f(\mathbf{X}^l, \mathbf{X}^r)$ from real-world stereo infrared (IR) images $(\mathbf{X}^l, \mathbf{X}^r)$. Given $K$ pairs of stereo images $\{(\mathbf{X}^l_i, \mathbf{X}^r_i )\}_{i = 1}^{K}$, $N$ pixels need to be selected whose ground truth depth will be obtained automatically by probing with tactile feedback, forming the touches $\mathbb{T}=\{t_1, t_2, \cdots, t_N\}$. Then $f$ is finetuned to $f^*$ with the collected depth labels. In our work, we use the pre-trained model from ActiveZero~\cite{liu2022activezero}, which demonstrates high performance on real IR images. However, our method can be applied to any cost-volume-based stereo network.

Worth to mention, finetuning a model until convergence takes $10 \sim 20$ minutes, while each probing in the real world takes $20 \sim 30$ seconds. It is not cost-effective for us to touch once and then finetune once. Therefore, it is preferred to set the number of touches $N>1$ before finetuning for efficiency. Specifically, we set $N=K\times n$ where $K$ corresponds to the number of stereo image pairs and $n$ is a pre-set average number of pixels to touch in each pair.

\subsection{Data Collection with Tactile Sensor (Stage 1)}
\label{sec:utility_def}

Since the data collection time is determined by the number of touches, we aim to improve the data efficiency by choosing the optimal set of touches.
We formalize the effect of touches $\mathbb{T}$ on the model as a utility function $U$, which indicates how much the stereo network’s performance improves given such touches.  We use $E_O$ to represent the depth prediction error on an object $O$, and $f^*_\mathbb{T}$ to represent the optimal model we can achieve through finetuning on the baseline model $f$ with touches $\mathbb{T}$. The utility function $U$ should measure the improvement of the model over all objects after finetuning based upon the touches $\mathbb{T}$, or equivalently to minimize the overall error of the finetuned model $f^*_\mathbb{T}$ since baseline model $f$ is fixed. 
\vspace*{-0.8\baselineskip}

\begin{align}
\nonumber U (\mathbb{T}) = - \sum_{o} E_o(f^*_\mathbb{T}) 
\end{align}

The optimal touching $\mathbb{T}^*$ is then chosen to maximize the utility function $U$.
\vspace*{-0.8\baselineskip}

\begin{align}
\nonumber \mathbb{T}^* &= \underset{\mathbb{T}}{\text{argmax}}~ U (\mathbb{T}) 
\end{align}

Therefore, the estimation of $U (\mathbb{T})$ is the prerequisite of the preference of $\mathbb{T}^*$.
However, $ U (\mathbb{T}) $ cannot be calculated directly for two reasons. Firstly, evaluating depth prediction for all objects ($O$) is impractical due to the absence of ground truth depth on untouched objects, making accurate error calculation impossible. Secondly, The utility function necessitates a depth label ($f^*_\mathbb{T}$) for finetuning. However, obtaining depth labels for touches depends on the utility function, creating a chicken-and-egg problem. Without estimating the utility function, obtaining depth labels is not feasible.

We propose to approximate $U (\mathbb{T})$ in two ways to overcome the above issues: approximating the improvement of depth prediction through confidence-based masks and approximating touch-based supervised finetuning through entropy-based self-supervised tuning.

\begin{figure}[t]
    \centering
    \begin{subfigure}[b]{0.49\linewidth}
    \centering
    \includegraphics[height=2cm]{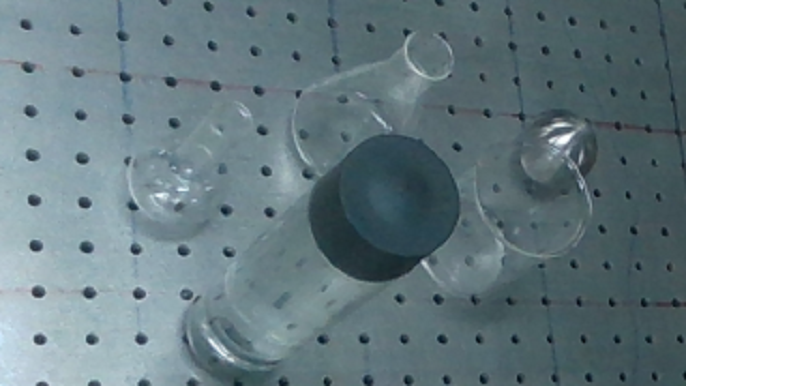}
    \caption{}
    \end{subfigure}
    \hfill
   \begin{subfigure}[b]{0.49\linewidth}
    \centering
    \includegraphics[height=2cm]{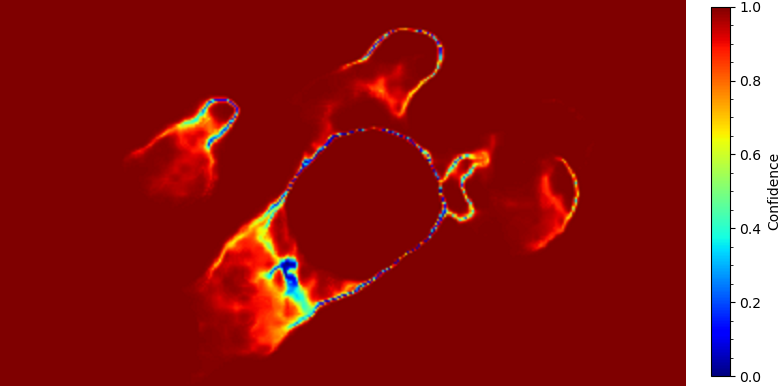}
     \caption{}
    \end{subfigure}
       \begin{subfigure}[b]{0.49\linewidth}
    \centering
    \includegraphics[height=2cm]{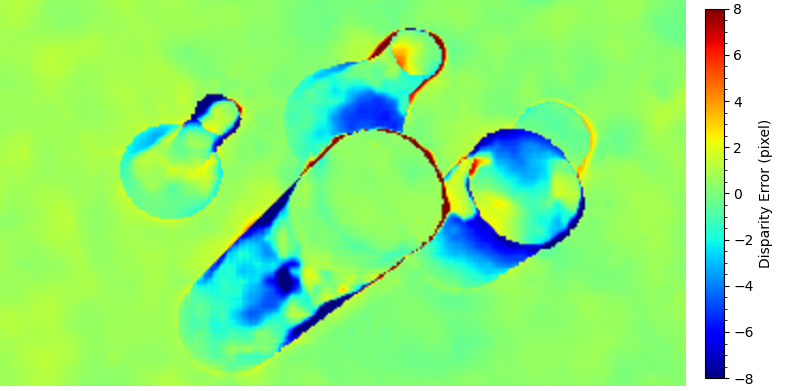}
     \caption{}
    \end{subfigure}
    \hfill
       \begin{subfigure}[b]{0.49\linewidth}
    \centering
    \includegraphics[height=2cm]{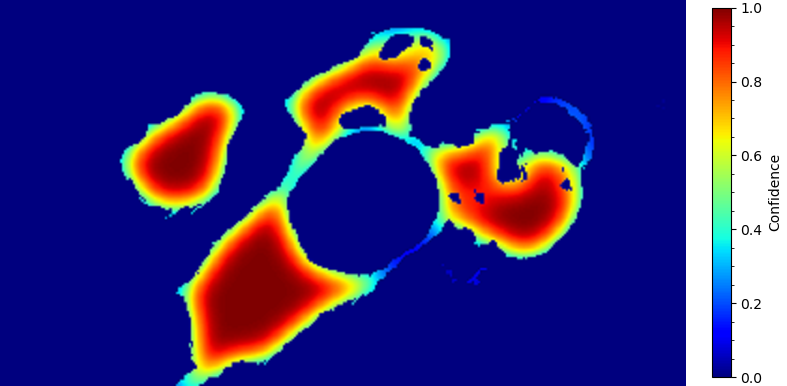}
     \caption{}
    \end{subfigure}
    \caption{Visualization of confidence-based mask: (a) RGB image; (b) confidence function  $\mathbf{C}_{f^*_{\mathbb{T}}}$; (c) pre-trained model's disparity error; (d) smoothed mask $G_U \circ M_{f^*_{\mathbb{T}}}$.
    }
    \label{fig: confidenceMask}
    \vspace*{-1.0\baselineskip}
\end{figure}

\noindent\textbf{Confidence-based Masks}
Ideally, the utility function should quantify the improvement of the model $f^*_{\mathbb{T}}$ with the reduction in depth prediction errors. However, without ground truth depth labels, such improvement is intractable in practice. 
Alternatively, we propose to use the reduction of unconfident areas to assess the improvement of the model. 

For our cost-volume-based stereo network $f^*_{\mathbb{T}}$, probability $p_{f^*_{\mathbb{T}}, (u, v)}$ over a pre-defined set of disparity hypotheses $\mathbb{D_H}$ is predicted for each pixel $(u, v)$, then the expected value of the disparity hypotheses is used as predicted disparity.
\vspace*{-0.8\baselineskip}

\begin{align}
    \nonumber f^*_{\mathbb{T}}(u, v) = \sum_{d_H \in \mathbb{D_H}} d_H \cdot p_{f^*_{\mathbb{T}}, (u, v)}(d_H)
\end{align}

The probability distribution $p_{f^*_{\mathbb{T}}, (u, v)}$ can be used to measure the depth estimation confidence~\cite{yao2018mvsnet,cheng2020deep}. Our confidence function $\mathbf{C}_{f^*_{\mathbb{T}}}$ is defined as a summation of the probability of the predicted disparity's neighborhood within a range of $\epsilon$.

\vspace*{-0.8\baselineskip}

\begin{align}
    \nonumber\mathbf{C}_{f^*_{\mathbb{T}}}(u, v) =\sum_{d_H \in \mathbb{D_H}} p_{f^*_{\mathbb{T}}, (u, v)} (d_H) \mathds{1}(|d_H - f^*_{\mathbb{T}}(u, v) | < \epsilon)
\end{align}

Summing the confidence over all the pixels gets the approximate utility function $U_{\text{confidence}}$:
\vspace*{-0.8\baselineskip}

\begin{align}
    \nonumber U_{\text{confidence}}(\mathbb{T}) 
 = \sum_{u, v} \mathbf{C}_{f^*_{\mathbb{T}}}(u, v)
\end{align}

The above utility function is inadequate primarily due to the presence of object boundaries. We characterize object boundaries by a sharp change in depth, which usually leads to a bimodal probability distribution of predicted depths. The confidence values near object boundaries are significantly lower compared to those in inaccurate prediction regions. However, our goal is to use confidence values to identify erroneous prediction areas, not boundaries. Thus, we binarize confidence values using a threshold $c_1$ to eliminate misleading low values near object boundaries. Then the mask of unconfident areas $\mathbf{M}_{f^*_{\mathbb{T}}}$ is computed as:
\vspace*{-0.8\baselineskip}

\begin{align}
    \nonumber\mathbf{M}_{f^*_{\mathbb{T}}}(u, v) = \left \{
        \begin{aligned}
            1 & \text{, if $\mathbf{C}_{f^*_{\mathbb{T}}}(u, v) \leq c_1$} \\
            0 & \text{, if $\mathbf{C}_{f^*_{\mathbb{T}}}(u, v) > c_1$}\\
        \end{aligned}
        \right.
\end{align}

where $c_1$ is the confidence threshold.
In order to further lower the weight of boundary pixels for subsequent optimization, we further apply a Gaussian Kernel $G_U$ on $\mathbf{M}_{f^*_{\mathbb{T}}}$ to form the approximated utility function $U$ as the weighted sum of all the unconfident areas:
\vspace*{-0.8\baselineskip}

\begin{align}
    \nonumber U_{\text{binarized}}(\mathbb{T}) = -\sum_{u, v}{(G_U \circ \mathbf{M}_{f^*_{\mathbb{T}}})(u, v)}
\end{align}

Fig.~\ref{fig: confidenceMask} shows an example of $\mathbf{C}_{f^*_{\mathbb{T}}}$, disparity estimation error, and $G_U \circ M_{f^*_{\mathbb{T}}}$ of transparent objects. As shown, the unconfident mask usually corresponds to regions with large disparity estimation errors.

\noindent\textbf{Surrogate Model with Entropy-based Self-supervised Tuning}

Having resolved the initial problem of lacking ground truth depth, we now face a second challenge - the presence of a chicken-and-egg dilemma. The expression for $U_{\text{binarized}}$ still relies on $f^*_{\mathbb{T}}$, which, unfortunately, remains unattainable during the data collection stage. This is because acquiring $f^*_{\mathbb{T}}$ necessitates ground truth depth, which is only available after fine-tuning.

As an alternative to supervised finetuning which requires the ground truth depth corresponding to touches $\mathbb{T}$, we use the entropy of the disparity distribution at pixels to touch as a self-supervised loss for tuning our model. In this way, we are actually using a surrogate model $f^{s}_{\mathbb{T}}$ trained with self-supervised loss to approximate the real finetuned model $f^*_{\mathbb{T}}$ trained with ground truth depth.
\vspace*{-0.8\baselineskip}

\begin{align}
    \nonumber\pazocal{L}_\text{entropy} = -  \sum_{t_{(u, v)}\in \mathbb{T}  } \sum_{d_H \in \mathbb{D_H}}  p_{f^{s}_{\mathbb{T}}, (u, v)} (d_H) \text{log} (p_{f^{s}_{\mathbb{T}}, (u, v)} (d_H))
\end{align}

$\pazocal{L}_\text{entropy}$ is aimed at minimizing the entropy of predicted probability along the disparity hypothesis for pixels to touch $\{(u, v) : t_{u, v} \in \mathbb{T}\}$. We found it a good surrogate for supervised finetuning in practice. By tuning the model to generate more confident predictions on pixels to touch, we can effectively identify the pixels that will be affected. An extra $L_2$ regularization on the disparity prediction is also introduced in practise. Pre-trained model $f$ is finetuned until convergence to $f^{s}_{\mathbb{T}}$. We then approximate the behavior of supervised finetuned model $f_{\mathbb{T}}^*$ with the self-supervised finetuned model $f_{\mathbb{T}}$ in the binarized utility function.
\vspace*{-0.8\baselineskip}

\begin{align}
    \nonumber U_{\text{surrogate}}(\mathbb{T}) = -\sum_{u, v}{(G_U \circ \mathbf{M}_{f^{s}_{\mathbb{T}}})(u, v)}
\end{align}

Up to now, we have finally transformed the intractable utility function $U(\mathbb{T})$ into a tractable one $U_{\text{surrogate}}(\mathbb{T})$.

\noindent\textbf{Greedy-based Utility Optimization}

The above two approximations make the utility function tractable. We can then select pixels to touch through maximizing the approximated utility function. However, even with the approximated utility function, solving the optimization problem is still NP-hard. We therefore design a greedy optimization scheme as an approximation, where we progressively grow the touch set. Each time a touch set with size $n$ is chosen out of a pair of stereo images. Combining all the touch sets from $K$ different images forms the overall touch set $\mathbb{T} $.

\begin{algorithm}
    \renewcommand{\algorithmicrequire}{\textbf{Input:}}
    \renewcommand{\algorithmicensure}{\textbf{Output:}}
    \caption{Pixel-choosing algorithm}
    \label{alg: algorithm-label}
    \begin{algorithmic}[1]
    \REQUIRE Pre-trained model $f$, IR images $\mathbf{X}_1, \mathbf{X}_2, \cdots, \mathbf{X}_K$
    \STATE Initialize $f_0^1 = f$ and $\mathbb{T} = []$
    \FOR {$k=1$ to $K$}
    {
        \FOR {$i=1$ to $n$}
        {   
            \STATE Compute smoothed mask $G_U \circ \mathbf{M}_{f_{i - 1}^k}$ on $\mathbf{X}_k$
            \STATE Choose $(u_i^k, v_i^k) = {\text{argmax}_{(u^k,v^k)}} G_U \circ \mathbf{M}_{f_{i - 1}^k} (u^k, v^k)$
            \STATE Add $t_i^k$ at $(u_i^k, v_i^k)$ to $\mathbb{T}$
            \STATE Update model $f_{i -1 }^k$ with entropy-based self-supervised tuning to $f_i^k$
        }
        \ENDFOR
        \STATE Set $f_{0}^{k+1} = f_{n}^{k}$
    }
    \ENDFOR
    \ENSURE The set of pixels to touch $\mathbb{T}$
    \end{algorithmic}
\end{algorithm}

Algorithm \ref{alg: algorithm-label} outlines our pixel-choosing strategy, which sequentially selects a new pixel based on the dynamic behavior of our stereo model. Initially, we use the pre-trained model $f_0^1 = f$ and calculate the smoothed mask of unconfident areas $G_U \circ \mathbf{M}f$. The first pixel for touch, $t_1^1$ at $(u_1^1, v_1^1) = {\text{argmax}_{(u^1,v^1)}} G_U \circ \mathbf{M}f (u^1, v^1)$, is then chosen. Subsequently, we update the model to $f_{i}^{j}$ with self-supervised entropy loss after selecting the $i$th pixel for touch of $j$th stereo images, denoted as $t_{i}^{j}$. The pre-trained model $f$ is finetuned with entropy loss at $(u_1^1, v_1^1)$ until convergence to $f_1^1$. For choosing the next pixel, we use $f_1^1$ to generate the smoothed mask of unconfident areas, ensuring avoidance of pixels with similar features due to their incorporation into $f_1^1$ during entropy-based self-supervised tuning. This approach reduces redundant sampling, making the greedy pixel-choosing algorithm implicitly maximize the surrogate utility function, as it consistently selects the least confident areas for touch, leading to increased confidence after finetuning.

Once the touching pixels are selected, ground truth depth labels of these pixels are acquired through our visual-tactile depth probing system, which will be described in Sec.~\ref{section: probing system}.

\subsection{Stereo Network Finetuning (Stage 2)}
\label{sec:tuning}

The finetuning method of the stereo network using the sparse depth labels has a significant impact on the network's performance. The depth labels are sparse and  only cover a few pixels in the stereo IR images, which poses a challenge for finetuning. Moreover, the network may suffer from catastrophic forgetting, where its performance on untouched regions is deteriorated after finetuning.
To tackle these two difficulties, we propose a combination of tactile loss and regularization loss.

\noindent\textbf{Tactile Loss} Upon each probing, we can obtain the ground truth depth $d(u, v)$ of pixel $(u, v)$ in the image. Because we have excluded object boundaries, we form a smooth approximation of the collected sparse depth label based on the assumption of continuity of object surfaces.

 For a $(2p + 1) \times (2p + 1)$ patch near the touch pixel, we first obtain a prior guess of depth from the pre-trained model's prediction, denoted as $\Tilde{d}$, then refine it with the collected depth $d(u, v)$. We use a Gaussian kernel $G_T$ to balance the weights of the prior guess and collected depth to get a refined depth of the small patch $\mathbf{P}(u, v)$.
\vspace*{-0.8\baselineskip}

\begin{align}
    \nonumber \pazocal{L}_{\text{tactile}} = \sum_{(u_p, v_p) \in \mathbf{P}(u,v)} \text{Smooth}L_{1}(f(u_p, v_p), d_{\text{refined}}(u_p, v_p))
\end{align}
\vspace*{-2.0\baselineskip}

\begin{align}
    \nonumber d_{\text{refined}}(u_p, v_p) = G_T(u_p, v_p) d(u, v) + (1 - G_T(u_p, v_p)) \Tilde{d}(u_p, v_p )
\end{align}
\vspace*{-2.0\baselineskip}

\begin{align}
    \nonumber G_T(u_p, v_p) = \exp ( {- \frac{(u_p - u)^2 + (v_p - v)^2}{2 \sigma_{T}^2}})
\end{align}
where $\mathbf{P}(u,v)$ denotes the $(2p + 1) \times (2p + 1)$ patch centering in pixel $(u, v)$. 
For center pixels, the refined depth is closer to touched pixels, since a small distance indicates depth is also close by continuity assumption. While for surrounding pixels, the refined depth is closer to depth prior $\Tilde{d}$, as the depth at the center may not be close to the actual depth here.

\noindent\textbf{Regularization Loss} 
Finetuning only with tactile loss meets with the catastrophic forgetting problem. We address this problem with a regularization loss, which computes the difference between the disparity prediction and the pre-trained model's prediction in confident areas.

For the pre-trained model used in our work, it has a small disparity error and high confidence on diffuse and specular objects. Therefore, we leverage disparity predictions from pre-trained model $f$ as pseudo depth label and use confidence to compute the pseudo label mask $\mathbf{M}_P$:
\vspace*{-0.8\baselineskip}

\begin{align}
    \nonumber \mathbf{M}_P(u,v) = \left \{
        \begin{aligned}
            1 & \text{, if $\mathbf{C}_{f^*(u, v)} \geq c_2$} \\
            0 & \text{, if $\mathbf{C}_{f^*(u, v)} < c_2$}\\
        \end{aligned}
        \right.
\end{align}
where $c_2$ is a pre-set confidence threshold. Pseudo loss is calculated by smooth $L_1$ loss:
\vspace*{-0.8\baselineskip}

\begin{align}
    \nonumber \pazocal{L}_{\text{regularization}} = \sum  
         \mathbf{M}_P \cdot \text{Smooth}L_{1} (
        f^*(\mathbf{X}^{l}, \mathbf{X}^r),f(\mathbf{X}^{l}, \mathbf{X}^r)
        )
\end{align}

\noindent\textbf{Combination}
The final combined loss for stereo network finetuning is defined as:
\vspace*{-0.8\baselineskip}

\begin{align}
    \nonumber \pazocal{L}_{\text{finetuning}} = \lambda_\text{T} \pazocal{L}_{\text{tactile}} + \lambda_\text{R} \pazocal{L}_{\text{regularization}}
\end{align}
where $\lambda_\text{T}$ and $\lambda_R$ are the weights for the tactile loss and regularization loss. Note that $f_n^k$ is not used in stereo network finetuning and we use $\pazocal{L}_{\text{finetuning}}$ to finetune the pretrained model $f$  to obtain the final finetuned model $f^*$.

\section{EXPERIMENTS}
\subsection{Visual-Tactile Depth Probing System}
\label{section: probing system}

\begin{figure}[t]
    \centering
    \includegraphics[width=5cm, angle=0]{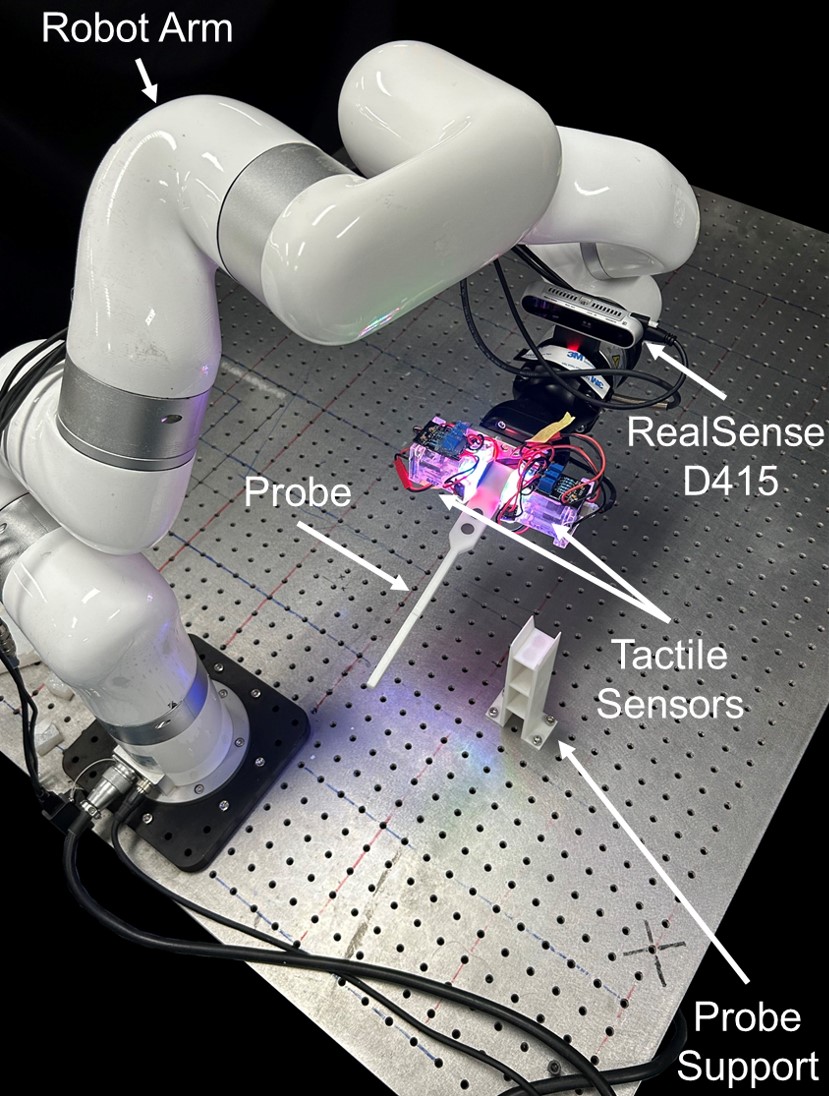}
    \caption{Visual-tactile depth probing system setup}
    \label{fig: probing}
    \vspace*{-1.0\baselineskip}

\end{figure}

Fig.~\ref{fig: probing} shows our visual-tactile depth probing system. It consists of a 7-DoF Xarm7 robot arm, a Robotiq Hand-E gripper, two CMOS-based tactile sensors, and a 3D printed probe with a length of 10 cm and a diameter of 5 mm. 
To ensure that the contact between the tactile sensor and the scene occurs only at the gel part of the sensor and not at any other parts, a probe is used for touching instead of applying the tactile sensor directly. This method avoids the need for a precise normal direction when touching with tactile sensors, which is difficult to estimate for transparent objects.

Before each probing, the robot grasps the probe from the support where it is attached by two magnets and moves it to a predefined initial position. Given the pixel that needs to be touched, since the ground truth depth of this pixel is unknown, We utilize the MoveIt ROS package as the robot's controller and control the probe to move along the camera ray direction of the pixel. For observing touch, we employ CMOS-based tactile sensors that are capable of detecting the displacement of the probe. The ground truth depth is obtained by recording the position of the robot arm's end effector when a touch event occurs. Our system's relocation error between multiple touches is within $3$ mm, which is negligible compared to the depth prediction error. It's important to note that we secure objects to the table using glue to prevent any unintended position changes during touch experiments.

\subsection{Evaluation Dataset}

\begin{figure}[t]
\centering
\includegraphics[width=0.98\linewidth]{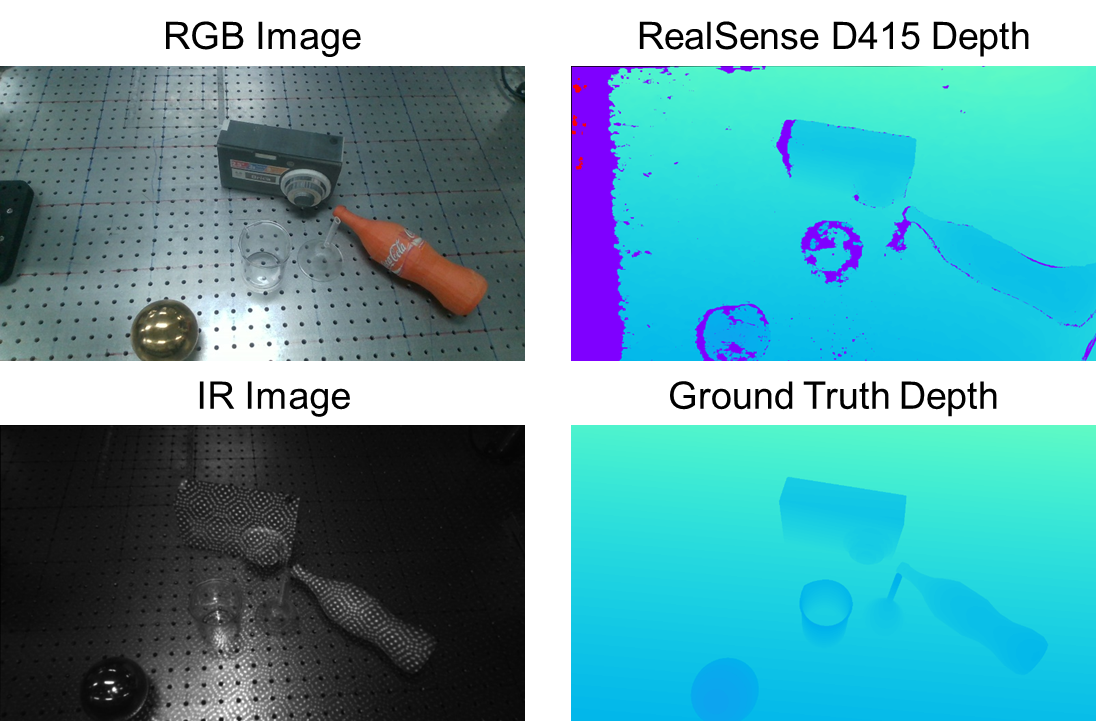}
\caption{An example of our real-world evaluation dataset}
\label{fig:dataset}
\vspace*{-1.0\baselineskip}

\end{figure}

To evaluate the effectiveness of the proposed method, we follow the pipeline in \cite{zhang2023close} to build a real-world evaluation dataset with ground truth depth as shown in Fig.~\ref{fig:dataset}. There are some improvements of our dataset against the one in~\cite{zhang2023close}: we improve the camera calibration accuracy; we modify some objects' CAD models to better match the real objects; we include $8$ fully transparent objects, $5$ of them have never even been seen before. Overall, the testing dataset consists of $550$ views in $50$ different scenes.

\subsection{Experiment Details}
We use the model from ActiveZero~\cite{liu2022activezero} as the pre-trained model. In \cite{liu2022activezero}, PSMNet~\cite{chang2018pyramid} is used as the stereo network's backbone, which extracts image features at different scales, forms a cost volume, and uses 3D CNN to regress the disparity. The disparity range is set to $12$ to $96$ pixels. 

For the depth probing, we build $5$ scenes by randomly placing objects on the table. For each scene, we acquire $4$ pairs of IR images from different views and select $n=5$ pixels to touch from each view using the touching utility function approximation. Therefore, we have $N=5\times4\times5=100$ pixels for all the experiments. Note that the object used in depth probing is a subset of the evaluation dataset, enabling us to test our method's generalizability. 

During our computation and optimization for our utility function, we first compute the confidence mask with neighborhood range $\epsilon = 5$. Then we clip the confidence mask with threshold $c_1 = 0.999$ and apply a Gaussian blur with $\sigma_U = 6.5$. The $L_2$ regularization loss weight is set as $\lambda_{L_2}=0.01$. We use Adam optimizer with a learning rate of $1e-5$.

For finetuning, we use Adam optimizer with a learning rate of $2e-5$, and it takes 10 epochs to converge. The network is finetuned on an NVIDIA RTX 3090 GPU, taking a total of $15$ minutes. We set $\lambda_\text{T} = 1.0$ and $\lambda_\text{R} = 100.0$ to make these two losses to a similar scale. Also, $p$ is set to $7$, $\sigma_T=12.0$ for the Gaussian kernel. $c_2=0.9999$ for the choice of pseudo depth labels. 

\subsection{Evaluation Metrics}

Common stereo estimation metrics are used to evaluate our results. End-point-error (EPE) is the mean absolute error of disparity. \textit{Bad1} is the percentage of pixels with disparity error larger than $1$. Absolute depth error (abs depth err) is calculated by converting the disparity to depth with camera parameters. In addition, we compute the percentage of pixels with absolute depth error larger than 4 mm ($>$4 mm), and the percentage of pixels with relative depth error smaller than $5\%$ ($\delta_{1.05}$). To evaluate our method for transparent objects, we compute all the metrics for both all objects and for transparent objects using the object label images.

\subsection{Experimental Results}

\begin{figure*}[ht]
    \centering
    \begin{subfigure}[b]{5.5cm}
    \centering
    \includegraphics[height=2.75cm]{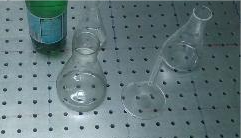}
    \caption{}
    \end{subfigure}
        \begin{subfigure}[b]{5.5cm}
    \centering
    \includegraphics[height=2.75cm]{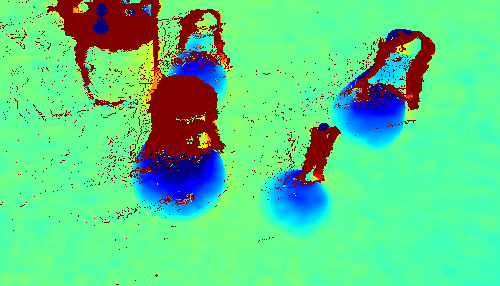}
    \caption{}
    \end{subfigure}
        \begin{subfigure}[b]{5.5cm}
    \centering
    \includegraphics[height=2.75cm]{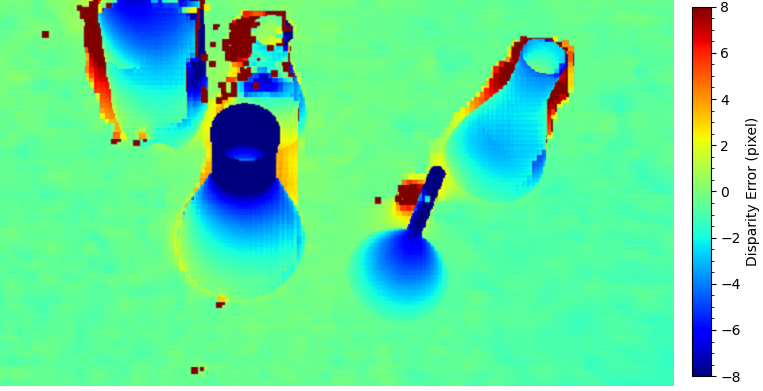}
    \caption{}
    \end{subfigure}
        \begin{subfigure}[b]{5.5cm}
    \centering
    \includegraphics[height=2.75cm]{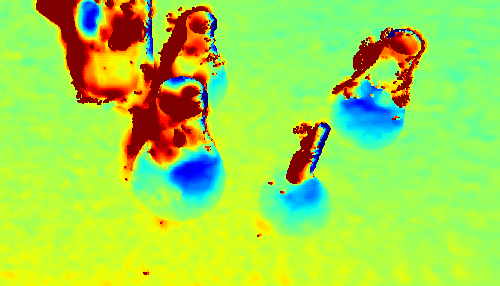}
    \caption{}
    \end{subfigure}
    \begin{subfigure}[b]{5.5cm}
    \centering
    \includegraphics[height=2.75cm]{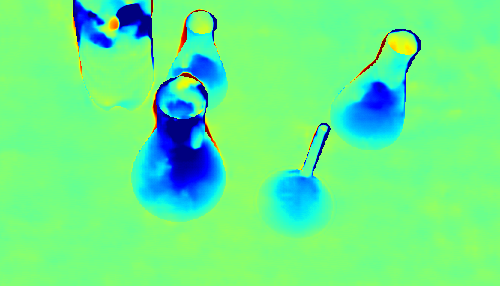}
    \caption{}
    \end{subfigure}
        \begin{subfigure}[b]{5.5cm}
    \centering
    \includegraphics[height=2.75cm]{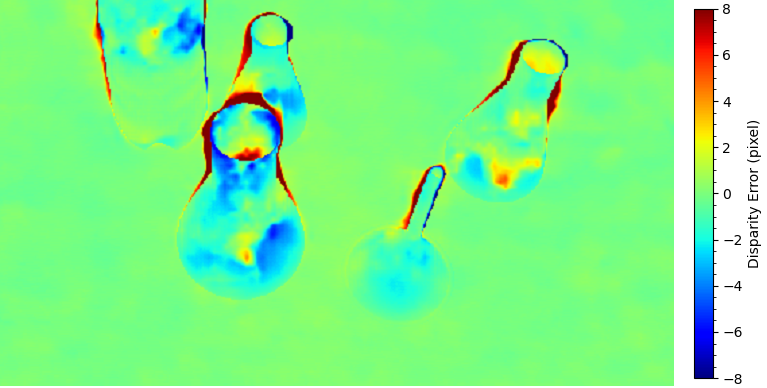}
    \caption{}
    \end{subfigure}
    \caption{Visualization of disparity error of different methods: (a) RGB image, (b) RealSense D415, (c) ClearGrasp, (d) TransCG, (e) ActiveZero, (f) Ours.}
    \label{fig: comparisonBetweenMethod}
\end{figure*}

\begin{table*}[ht]
\caption{Comparison of disparity estimation on the real-world evaluation dataset.}
    \label{table: method}
    \centering
    \begin{tabular}{c|cc|cc|cc|cc|cc}
    \hline
    \multirow{2}{*}{Method} & \multicolumn{2}{c|}{EPE(px) $\downarrow$}                & \multicolumn{2}{c|}{Bad1 $\downarrow$}              & \multicolumn{2}{c|}{Abs Depth Err (mm) $\downarrow$}          & \multicolumn{2}{c|}{\textgreater{}4 mm $\downarrow$}      & \multicolumn{2}{c}{$\delta_{1.05} \uparrow$}              \\ \cline{2-11} 
                            & \multicolumn{1}{c|}{All}      & Trans & \multicolumn{1}{c|}{All} & Trans & \multicolumn{1}{c|}{All}      & Trans & \multicolumn{1}{c|}{All}      & Trans & \multicolumn{1}{c|}{All}      & Trans \\ \hline
    RealSense D415               & \multicolumn{1}{c|}{3.290} & 4.371    & \multicolumn{1}{c|}{0.639}    &  0.819          & \multicolumn{1}{c|}{19.805} & 25.160    & \multicolumn{1}{c|}{0.832} & 0.929    & \multicolumn{1}{c|}{0.768} & 0.671    \\ \hline
    ClearGrasp              & \multicolumn{1}{c|}{3.323} & 3.786    & \multicolumn{1}{c|}{0.749}    &    0.831     & \multicolumn{1}{c|}{22.727} & 24.983    & \multicolumn{1}{c|}{0.900} & 0.942    & \multicolumn{1}{c|}{0.736} & 0.697    \\ \hline
    TransCG                 & \multicolumn{1}{c|}{4.424} & 4.461    & \multicolumn{1}{c|}{0.873}    & 0.852     & \multicolumn{1}{c|}{26.146} & 25.647    & \multicolumn{1}{c|}{0.956} & 0.949    & \multicolumn{1}{c|}{0.694} & 0.699    \\ \hline
    ActiveZero 
    (pre-trained model)             & \multicolumn{1}{c|}{1.424} & 2.083    & \multicolumn{1}{c|}{0.395}    &   0.605    & \multicolumn{1}{c|}{12.332} & 17.562    & \multicolumn{1}{c|}{0.648} & 0.831    & \multicolumn{1}{c|}{0.906} & 0.846    \\ \hline
    Ours                    & \multicolumn{1}{c|}{\textbf{1.296}} & \textbf{1.875}    & \multicolumn{1}{c|}{\textbf{0.354}}  & \textbf{0.541}        & \multicolumn{1}{c|}{\textbf{11.123}} & \textbf{15.591}    & \multicolumn{1}{c|}{\textbf{0.628}} & \textbf{0.799}    & \multicolumn{1}{c|}{\textbf{0.932}} & \textbf{0.890}    \\ \hline
    \end{tabular}
\end{table*}

\begin{table*}[ht]
\caption{Comparison of different pixel selection strategy}
    \centering
    \begin{tabular}{c|cc|cc|cc|cc|cc}
    \hline
    \multirow{2}{*}{Strategy} & \multicolumn{2}{c|}{EPE(px) $\downarrow$}                & \multicolumn{2}{c|}{Bad1 $\downarrow$}              & \multicolumn{2}{c|}{Abs Depth Err (mm) $\downarrow$}          & \multicolumn{2}{c|}{\textgreater{}4 mm $\downarrow$}      & \multicolumn{2}{c}{$\delta_{1.05} \uparrow$}              \\ \cline{2-11} 
                            & \multicolumn{1}{c|}{All}      & Trans & \multicolumn{1}{c|}{All} & Trans & \multicolumn{1}{c|}{All}      & Trans & \multicolumn{1}{c|}{All}      & Trans & \multicolumn{1}{c|}{All}      & Trans \\ \hline
    Random                    & \multicolumn{1}{c|}{1.378} & 2.014    & \multicolumn{1}{c|}{0.380}    &  0.585           & \multicolumn{1}{c|}{11.893} & 16.917    & \multicolumn{1}{c|}{0.642} & 0.822    & \multicolumn{1}{c|}{0.917} & 0.864    \\
    Confidence                & \multicolumn{1}{c|}{1.383} & 2.013    & \multicolumn{1}{c|}{0.384}    &  0.586           & \multicolumn{1}{c|}{11.954} & 16.920    & \multicolumn{1}{c|}{0.644} & 0.824    & \multicolumn{1}{c|}{0.914} & 0.860    \\ 
    Ours                      & \multicolumn{1}{c|}{\textbf{1.296}} & \textbf{1.875}    & \multicolumn{1}{c|}{\textbf{0.354}}    & \textbf{0.541}            & \multicolumn{1}{c|}{\textbf{11.123}} & \textbf{15.591}    & \multicolumn{1}{c|}{\textbf{0.628}} & \textbf{0.799}    & \multicolumn{1}{c|}{\textbf{0.932}} & \textbf{0.890}    \\ \hline
    Human                     & \multicolumn{1}{c|}{1.262} & 1.826    & \multicolumn{1}{c|}{0.332}    &   0.512    & \multicolumn{1}{c|}{10.689} & 14.940    & \multicolumn{1}{c|}{0.614} & 0.783    & \multicolumn{1}{c|}{0.933} & 0.893    \\ \hline
    \end{tabular}
    \label{table: strategy}
\vspace*{-1.0\baselineskip}

\end{table*}

We compare the performance of our finetuned model with a commercial depth sensor (Intel RealSense D415) and several depth estimation methods, including ClearGrasp~\cite{sajjan2020clear}, TransCG~\cite{fang2022transcg}, and our pre-trained model, Activezero~\cite{liu2022activezero}. 
ClearGrasp predicts a transparent object mask on the RGB image, and uses global optimization to complete the depth from the commercial depth sensor with normal and boundary constraints. 
TransCG employs a 2D network to impaint the incomplete depth, and the network is trained on a large-scale real-world dataset of transparent objects. ActiveZero is a learning-based stereo method that is trained using mixed domain learning. Note that we do not collect the ground truth depth for the whole image. Thus, ClearGrasp and TransCG can only be trained on their own datasets and may face domain-shifting problems. Comparison results are shown in Table~\ref{table: method} and Fig.~\ref{fig: comparisonBetweenMethod}. Our method outperforms other methods in all metrics. 
After tuning on the sparse touches, the average absolute depth error on transparent objects is reduced from 17.5 mm to 15.5 mm. Also, Bad1 metric is improved by around $10 \%$ compared with the pre-trained model, showing that we successfully remove more outliers in the model prediction than the sparse depth labels of 100 pixels.

\subsection{Ablation Study}
\noindent\textbf{Touch Selection} To illustrate that our utility function and optimization scheme improve sample efficiency, we compare our method with other pixel-choosing strategies, including randomly choosing a pixel in the image (\textit{Random}), choosing pixels with the lowest confidence (\textit{Confidence}), and manually selecting pixels for touch (\textit{Human}). When manually selecting pixels to touch, we pick the center pixel of transparent objects. To prevent redundant touching in the same area, the Confidence strategy is further restricted so that any two touches have a distance of at least 20 pixels. Table~\ref{table: strategy} compares the finetuned networks' disparity estimation accuracy.
It is shown that  our pixel-choosing strategy leads to a significant improvement in the finetuning process, compared to Random or Confidence strategies. Moreover, our strategy achieves a performance improvement that is comparable to the manual strategy.

\begin{table}[ht]
\caption{Ablation study on Gasussian on tactile loss}
\centering
    \begin{tabular}{c|c|c}
    \hline
    \multirow{2}{*}{Tactile loss} & \multicolumn{2}{c}{EPE(px)}             \\ \cline{2-3} 
                                  & {All}      & Trans    \\ \hline
    Pixel-wise                    & {1.372} & 2.019 \\ \hline
    Patch-wise                    & \textbf{1.296} & \textbf{1.875} \\ \hline
    \end{tabular}
    \label{table: gaussianSmoothing}
\end{table}

\begin{table}[ht]
\caption{Ablation study on regularization loss}
\centering
    \begin{tabular}{c|c|c}
    \hline
    \multirow{2}{*}{Loss}    & \multicolumn{2}{c}{EPE(px)}             \\ \cline{2-3} 
                             & {All}      & Trans    \\ \hline
    tactile                  & {1.348} & 1.961 \\ \hline
    tactile + regularization & \textbf{1.296} & \textbf{1.875} \\ \hline
    \end{tabular}
    \label{table: regularization loss}
    \vspace*{-1.0\baselineskip}

\end{table}

\noindent\textbf{Gaussian on tactile loss} In Sec.~\ref{sec:tuning}, we use a Gaussian kernel and prior depth guess to propagate the depth signal of a pixel to its neighborhood. Despite it is a biased estimation of the ground truth depth, Gaussian smoothing relieves the sparsity of tactile signals and improves the finetuned model's performance, as shown in Table~\ref{table: gaussianSmoothing}.

\noindent\textbf{Regularization Loss}
To examine how regularization loss affects finetuning, we experiment with different loss combinations as shown in Table~\ref{table: regularization loss}. We find that imposing a regularization loss on high-confidence regions also enhances the depth estimation of transparent objects. This suggests that the regularization loss facilitates the model to leverage depth cues more effectively and mitigate the ''catastrophic forgetting'' problem.

\section{CONCLUSION AND DISCUSSION}

In this paper, we have introduced a novel method aimed at enhancing real-world depth sensing for transparent objects by finetuning a stereo network with sparse depth labels automatically collected using a probing system with tactile feedback. We propose a novel utility function to evaluate touch benefits and improve the network's performance with a fixed touching budget by optimizing the utility function. We also use tactile depth supervision and confidence-based regularization during finetuning. To evaluate our approach, we constructed a dataset including diffuse and transparent objects. Experimental results illustrate the effectiveness of our method in improving depth sensing accuracy for transparent objects. 

A significant advantage of our method lies in its ability to operate without the need for a pre-collected dataset, as it actively selects touch points and enhances sample efficiency. This unique feature sets our approach apart and helps us overcome the challenge of labor-intensive data collection. Moreover, our work demonstrates the potential of integrating robotic interactions with the physical environment and utilizing proprioception to gather data. With this capability, the robots can autonomously and consistently enhance their perception methods in real-world scenarios.

However, our method does have some limitations that warrant consideration for future research. Firstly, the utility optimization based on a greedy strategy may not consistently capture the global optimum, suggesting room for further exploration of optimization techniques. Furthermore, the generalization ability of our method remains unexamined, prompting the need for extensive testing across diverse scenarios and object types.

Looking ahead, our method holds promising applications in real-world robotic systems equipped with tactile sensors, particularly in grasping and manipulating transparent objects. We anticipate that our work will serve as an inspiration for future research endeavors, spurring further exploration and adoption of our method in various real-world scenarios.

\bibliographystyle{IEEEtran}
\bibliography{main}
\end{document}